\documentclass[10pt,twocolumn,letterpaper]{article}

\usepackage{wacv}              

\usepackage{multirow}
\usepackage{booktabs}
\usepackage{diagbox}
\usepackage{arydshln} 
\usepackage{makecell}
\definecolor{wacvblue}{rgb}{0.21,0.49,0.74}
\usepackage[pagebackref,breaklinks,colorlinks,allcolors=wacvblue]{hyperref}

\def\wacvPaperID{579} 
\def\confName{WACV}
\def\confYear{2026}

\title{Language Integration in Fine-Tuning Multimodal Large Language Models for Image-Based Regression}

\author{Roy H. Jennings, Genady Paikin, Roy Shaul, Evgeny Soloveichik \\
Samsung Israel R\&D Center \\
Tel-Aviv, Israel \\
{\tt\small\{roy.jennings, genady.paikin, roy.shaul, evgeny.soloveichik\}@samsung.com}}

\begin{document}
\maketitle
\begin{abstract}
Multimodal Large Language Models (MLLMs) show promise for image-based regression tasks, but current approaches face key limitations. 
Recent methods fine-tune MLLMs using preset output vocabularies and generic task-level prompts (e.g., "How would you rate this image?"), assuming this mimics human rating behavior.
\textbf{Our analysis reveals that these approaches provide no benefit over image-only training}. Models using preset vocabularies and generic prompts perform equivalently to image-only models, failing to leverage semantic understanding from textual input.
We propose \textbf{Regression via Transformer-Based Classification} (RvTC), which replaces vocabulary-constrained classification with a flexible bin-based approach.
Unlike approaches that address discretization errors through complex distributional modeling, RvTC eliminates manual vocabulary crafting through straightforward bin increase, achieving state-of-the-art performance on four image assessment datasets using only images.
\textbf{More importantly, we demonstrate that data-specific prompts dramatically improve performance}. 
Unlike generic task descriptions, prompts containing semantic information about specific images enable MLLMs to leverage cross-modal understanding. 
On the AVA dataset, adding challenge titles to prompts substantially improves our already state-of-the-art image-only baseline. 
We demonstrate through empirical evidence from the AVA and AGIQA-3k datasets that MLLMs benefit from semantic prompt information, surpassing mere statistical biases. 
We validate RvTC across two different MLLM architectures, demonstrating consistent improvements and method generalizability.
\footnote{The source code is available at \url{https://github.com/royhj/rvtc}.}
\end{abstract}
    
\section{Introduction}
\label{sec:intro}
Vision-language models trained on massive unlabeled image-language datasets have demonstrated a remarkable capacity to extract universal image features, with CLIP \cite{radford2021clip} achieving impressive zero-shot classification performance on benchmark datasets such as ImageNet. 
Building on these foundations, Multimodal Large Language Models (MLLMs) have evolved to seamlessly fuse image and text embeddings with generative language capabilities \cite{kim2021vilt, liu2024llava, ye2024mplug}. 
This has sparked growing interest in transferring MLLM capabilities to image-based regression tasks, including Image Quality Assessment (IQA), Image Aesthetics Assessment (IAA) \cite{ke2023vila, wu2024qalign}, and AI-Generated Image Quality Assessment (AIGIQA) \cite{yang2024moe, peng2024aigc}.

\begin{figure*}
  \centering
    \includegraphics[width=1.\textwidth]{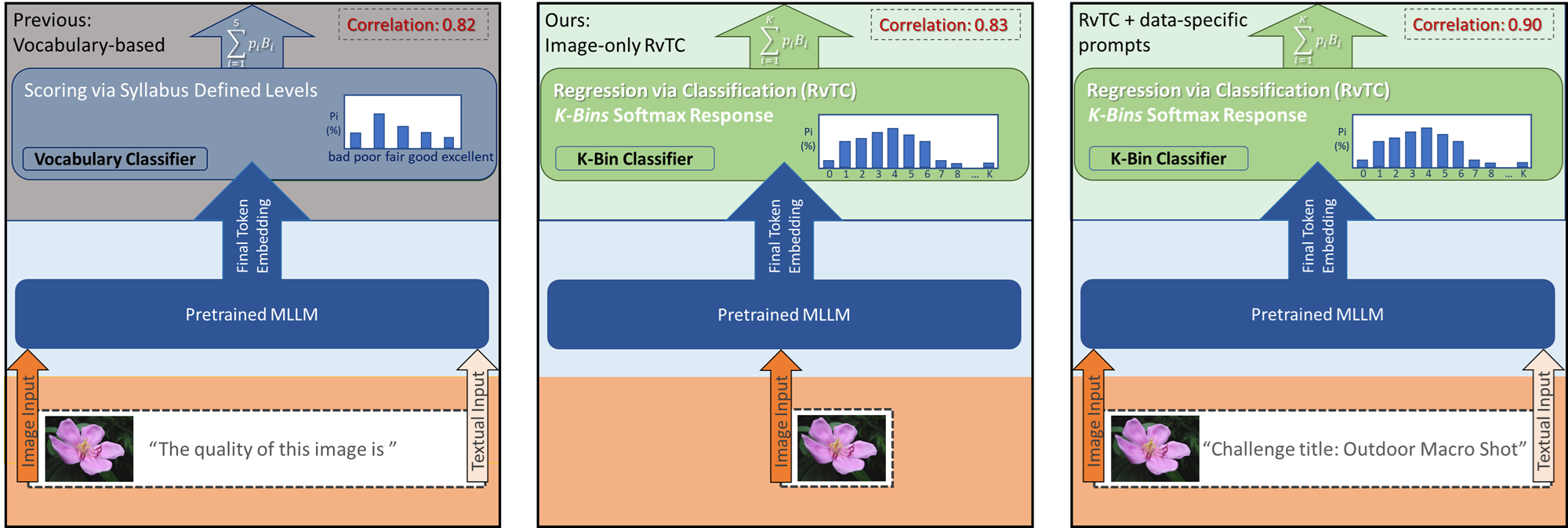}
\caption{\label{fig:outline}
Existing MLLM methods using preset vocabulary and generic prompts (left) achieve 0.82 correlation on AVA. 
Our image-only RvTC model (center) exceeds this with state-of-the-art correlation of 0.83. 
Integrating data-specific prompts (e.g., "Outdoor Macro Shot") during fine-tuning (right) unlocks MLLM cross-modal reasoning, yielding a new state-of-the-art 0.90.
}
\end{figure*}

Recent approaches \cite{lai2025font, lai2025snowmaster, wu2024qalign, wu2024qinstruct} fine-tune MLLMs for image regression using two key assumptions borrowed from human rating behavior: (1) utilizing preset output vocabularies (e.g., "excellent", "good", "fair", "poor", "bad"), and (2) incorporating generic task-level prompts such as "How would you rate the quality of this image?". 
These methods assume that mimicking human-like vocabulary and task descriptions will leverage the multimodal capabilities of MLLMs.

However, we demonstrate that current multimodal approaches provide no benefit over image-only training. 
Models using preset vocabularies and generic prompts perform equivalently to image-only models. 
This challenges the core assumption that current MLLM fine-tuning strategies effectively utilize cross-modal capabilities for regression tasks.
To address these limitations, we make three key contributions:

\begin{table}
    \centering
    \caption{\label{tab:ava_perf}Comparison of models on the AVA dataset.
    RvTC achieves state-of-the-art correlation of 0.83 without textual input, and rises to 0.90 with data-specific prompts (RvTC+).
    “LP” denotes linear probing of the regression head; “+” indicates inclusion of challenge titles.
    Results are reported using mPLUG-Owl2 as backbone (out baseline architecture); see \cref{tab:rvtc_cross_model} for validation with QWEN2-VL-2B, which yields even stronger results.}
    \begin{tabular}{l|ll}
        \toprule
        Method & SRCC & PLCC \\ 
        \midrule
        RvTC-LP & 0.709 & 0.711 \\
        RvTC-LP+ & 0.742 & 0.741 \\
        \midrule
        NIMA \cite{talebi2018nima} & 0.612 & 0.636 \\ 
        MUSIQ \cite{ke2021musiq} & 0.726 & 0.738 \\ 
        VILA \cite{ke2023vila} & 0.774 & 0.774 \\ 
        LIQE \cite{zhang2023liqe} & 0.776 & 0.763 \\ 
        One-Align \cite{wu2024qalign} & 0.823 & 0.819 \\ 
        \textbf{RvTC (ours)} & \textbf{0.833} & \textbf{0.831} \\
        \textbf{RvTC+ (ours)} & \textbf{0.899} & \textbf{0.901} \\
        \bottomrule
    \end{tabular}
\end{table}

\textbf{1. Rethinking regression with RvTC}: We propose \textit{Regression via Transformer-Based Classification} (RvTC), which replaces the rigid vocabulary constraints of previous methods with a flexible bin-based regression scheme. This simple yet effective approach outperforms prior multimodal methods using only images, matching or setting new state-of-the-art on four benchmarks.
Unlike recent work that addresses discretization errors through complex distributional modeling \cite{you2025teaching}, RvTC achieves superior accuracy by simply increasing the number of classification bins. 

\textbf{2. Enhancing MLLM performance through data-specific prompts}: 
We demonstrate that generic task prompts (e.g., "How would you rate this image?") fail to leverage cross-modal capabilities. 
In contrast, fine-tuning with data-specific prompts yields substantial improvements in correlation results. 
For example, the prominent IAA dataset AVA includes \textit{challenge titles} that characterize images with descriptive phrases such as "Rule of Thirds" or "Outdoor Macro Shot." 
Incorporating these semantic descriptors during fine-tuning substantially improves correlation results from our already state-of-the-art image-only baseline.
We validate this approach across two different MLLM architectures, demonstrating consistent improvements and method generalizability.

\textbf{3. Disentangling semantics from bias}: 
Through controlled experiments on AVA and AGIQA-3k datasets, we show that a significant portion of the observed gains stem from cross-modal understanding rather than dataset-specific statistical artifacts. 

Our findings demonstrate that effective fine-tuning of MLLMs for regression tasks requires moving beyond human-like vocabulary and generic prompts. 
Instead, training should incorporate semantically meaningful, data-specific context that aligns with the model’s cross-modal capabilities. 
When fine-tuned with such input, MLLMs exhibit significant gains in regression accuracy, revealing their potential for grounded visual understanding.

\section{Related work}
\textbf{REgression using CLAssification (RECLA)}.
RECLA transforms regression into classification by discretizing continuous targets into bins and training a classifier to predict the appropriate bin \cite{torgo1997regression}. 
Predictions are mapped back to continuous values through weighted averaging of bin centers. 
We show that RECLA integrates seamlessly with pre-trained MLLMs through straightforward fine-tuning.

\textbf{Image assessment regression tasks using MLLMs}.
Recent work has applied MLLMs such as mPLUG-Owl2 \cite{ye2024mplug} to Image Quality Assessment (IQA), Image Aesthetic Assessment (IAA), and AI-Generated Image Quality Assessment (AIGIQA). 
In IQA, the goal is to predict the perceived quality of an image, often by learning to map image features to subjective quality scores. 
In IAA, the objective is to assess the aesthetic appeal of an image, typically by predicting scores that reflect human aesthetic preferences.
In AIGIQA, generated images are evaluated for perceptual quality and alignment with the generation prompt.
The recent Q-Align \cite{wu2024qalign} teaches an MLLM for visual rating aligned with human opinions. 
Q-Align achieves state-of-the-art performance on IQA, IAA, and Video Quality Assessment (VQA) tasks. 
Q-Align unifies the three tasks into one model they call OneAlign.

Our work extends Q-Align by replacing vocabulary-based classification with flexible bin-based regression and demonstrates that semantically relevant textual prompts significantly improve performance. While concurrent work DeQA-Score \cite{you2025teaching} addresses discretization through soft label distributions, we show that simply increasing bin counts in a straightforward classification framework achieves superior results without complex distributional modeling.

\section{Method}
In this section, we present \textit{Regression via Transformer-Based Classification (RvTC)}, a framework that transforms multimodal regression into a classification problem with flexible bin counts. Unlike existing approaches that constrain outputs to preset vocabularies, RvTC eliminates manual vocabulary crafting while achieving superior performance through straightforward bin increase.

\subsection{Architecture overview}
We build our baseline model of RvTC upon the Multimodal Large Language Model mPLUG-Owl2 \cite{ye2024mplug}, which demonstrates strong visual perception and language understanding capabilities. The base architecture comprises: (1) Vision encoder: ViT-L/14 \cite{radford2021clip} processes input images. (2) Visual abstractor: Reduces visual features to 64 semantic token embeddings per image. (3) Language decoder: LLaMA-2-7B \cite{touvron2023llama} serves as a universal interface for mixed vision-language input.
We replace mPLUG-Owl2's vocabulary-constrained classification head with a $K$-bin linear classification head that supports arbitrary bin counts for regression tasks.
The bin classification head is applied to the penultimate hidden-state embedding of the final token, which acts as an aggregator for all of the tokens.

\subsection{Regression using classification framework}

\textbf{Problem formulation}.
Building on the RECLA (REgression using CLAssification) framework established by \cite{torgo1997regression}, RvTC reformulates regression problems $f: \mathbb{R}^d \rightarrow \mathbb{R}$ as classification problems $g: \mathbb{R}^d \rightarrow {1, 2, \ldots, K}$, where $K$ represents the number of bins. This transformation involves:
(1) Discretization: Target values are discretized into $K$ distinct bins using uniform binning over a preset min-max range.
(2) Assignment: Each target value is assigned to the bin whose center is closest to the original value.
(3) Classification: Inputs are classified using a linear head with $K$-bins as classes.

\subsection{Training and inference}
For training, we use standard cross-entropy loss to optimize bin classification.
During inference, posterior probabilities ${p_1, p_2, \ldots, p_K}$ are computed via softmax and then converted to continuous values through a weighted sum: $\sum_{i=1}^K p_i b_i$, where $b_i$ is the center of bin $i$.

We also evaluated ordinal regression \cite{niu2016ordinal} trained on the same K-bin discretization as thresholds for binary cross-entropy.
Across all datasets, ordinal regression matched standard cross-entropy within experimental noise and did not yield consistent improvements. For simplicity and reproducibility, we therefore use cross-entropy throughout.

\subsection{Advantages of the bin-based approach}
This formulation offers simplicity and flexibility over vocabulary-constrained methods; bin count can be adjusted without redefining vocabularies, and performance improves monotonically with increased number of bins (see \cref{fig:grid_rvtc}) in contrast to complex distributional modeling \cite{you2025teaching}.

\begin{figure}[t]
  \centering
  \begin{subfigure}{0.45\textwidth}
    \centering
    \includegraphics[width=\linewidth]{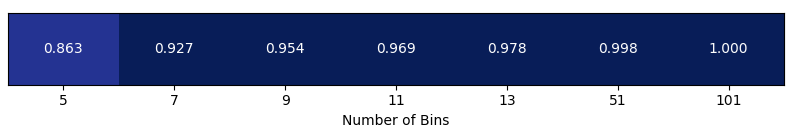}
  \end{subfigure}
  \caption{\label{fig:quant_noise}Effect of bin count on quantization noise, measured as (SRCC + PLCC)/2 between AVA ground truth mean opinion score values and their corresponding bin centers}
\end{figure}

Note that when quantizing the target values of the data, there is freedom in setting the quantization scheme; for example, one may use a non-uniform quantization scheme.
However, increasing the number of bins is a straightforward way to reduce quantization noise and simplify training, removing the need to tune hyperparameters of the range at the cost of added complexity to the classification task.
\cref{fig:quant_noise} quantifies this effect, demonstrating a near-perfect correlation between original targets and bin centers at 51+ bins.

\subsection{Training configurations}
\textbf{Image-only training}.
When fine-tuning without textual prompts, we refer to the model as \textit{image-only} RvTC. 
This configuration serves as our baseline and demonstrates that effective regression can be achieved using only visual features.

\textbf{Multimodal training with data-specific prompts}.
For multimodal training, we incorporate data-specific prompts that contain semantic information relevant to individual images rather than generic task descriptions. 
This approach, denoted RvTC+, enables the model to leverage cross-modal understanding for improved regression performance.

Throughout this paper, we use RvTC when referring to image-only training and full model fine-tuning; RvTC+ indicates inclusion of data-specific prompts with full fine-tuning, and the suffix "LP" denotes linear probing, where only the regression head is trained while the backbone remains frozen.

\section{Experiments}
\label{sec:empirical}

\subsection{Experimental setup}

\textbf{Datasets.} We conduct experiments on five datasets spanning different image assessment tasks. 
For \textbf{Image Aesthetic Assessment (IAA)}, we use AVA \cite{murray2012ava}, a prominent benchmark extracted from the DPChallenge website containing over 250,000 photographic images with mean opinion scores (MOS) ranging from 1 to 10. Each image belongs to one of 1,400 challenges with descriptive titles such as "Rule of Thirds" and "Self Portrait Without People"; these challenge titles serve as our data-specific prompts.

For \textbf{Image Quality Assessment (IQA)}, we evaluate on three datasets: KonIQ-10k \cite{hosu2020koniq} (10k images from the extensive YFCC100M multimedia database), SPAQ \cite{fang2020spaq} (11k smartphone photos), and KADID-10k \cite{lin2020kadid} (10k images with 25 distortion types at 5 intensity levels). For \textbf{AI-Generated Image Quality Assessment (AIGIQA)}, we use the AGIQA-3k dataset \cite{li2023agiqa}, containing 3k AI-generated images with corresponding generation prompts and MOS ratings for semantic alignment and perceptual quality.

All evaluations use official test sets with Spearman's Rank Correlation Coefficient (SRCC) and Pearson Linear Correlation Coefficient (PLCC) as metrics.

\textbf{Model architecture and training.} Our RvTC framework builds on mPLUG-Owl2 \cite{ye2024mplug}, using ViT-L/14 \cite{radford2021clip} as the vision encoder and LLaMA-2-7B \cite{touvron2023llama} as the language decoder. We replace mPLUG-Owl2's vocabulary classification head with a randomly initialized linear classification head for $K$-bin classification.

We fine-tune the entire model using Adam optimizer \cite{kingma2014adam} with cosine learning rate scheduling \cite{loshchilov2016sgdr}. 
The learning rate is linearly warmed up from 0 to 1e-5 over the first 3\% of total training steps.
For experiments without textual prompts, we use batch size 128 for 2 epochs; with textual prompts, we train for 3 epochs. All experiments use 4 NVIDIA RTX H100 GPUs.

Based on the results presented in \cref{fig:grid_rvtc}, we set the number of bins to 51 and use uniform binning over the preset min-max range of each training dataset, with target values replaced by the bins whose center is closest to the target value.

\textbf{Dataset-specific adaptations.} For AGIQA-3k, due to its smaller size, we first pre-fine-tune on AVA for 2 epochs before task-specific fine-tuning. This transfer learning approach leverages the larger aesthetic assessment dataset to improve performance on the AI-generated image task.
\subsection{Baseline performance: image-only RvTC}
\label{sec:task_level_prompt}
In this section, we establish RvTC's image-only baseline performance to isolate the contribution of textual prompts evaluated in \cref{sec:group_level_prompts}. 
Our RvTC approach achieves state-of-the-art results using only visual input across multiple datasets.

\textbf{Linear probing analysis}. \cref{tab:ava_perf} shows results for RvTC-LP (linear probing), where only the regression head is fine-tuned while the backbone remains frozen. 
This approach achieves strong image-only performance on AVA (SRCC: 0.709, PLCC: 0.711), demonstrating the generalization capabilities of mPLUG-Owl2's pre-trained visual representations without requiring full model fine-tuning (we refer to \mbox{RvTC-LP+} and \mbox{RvTC+} in \cref{sec:group_level_prompts}).

\begin{table}
    \centering
    \caption{\label{tab:iqa_perf_image_only}Performance comparison in SRCC/PLCC of image-only RvTC with different models on IQA tasks}
    \begin{tabular}{l|c|c|c}
        \toprule
         & KonIQ-10k & SPAQ & KADID-10k  \\
        \midrule
        NIMA \cite{talebi2018nima} & 0.859/0.896 &  0.907/0.910 & NA/NA \\ 
        MUSIQ \cite{ke2021musiq} & 0.916/0.928 & 0.917/0.921 & NA/NA \\ 
        LIQE \cite{zhang2023liqe} &  0.919/0.908 & 0.922/0.919 & 0.930/0.931 \\ 
        Q-Align \cite{wu2024qalign} & 0.940/0.941 & \textbf{0.930/0.933} & 0.919/0.918  \\ 
        \hdashline
        \textbf{RvTC (ours)} & \textbf{0.942/0.952} & 0.926/0.930 & \textbf{0.978/0.981} \\
        \bottomrule
    \end{tabular}
\end{table}

\begin{table}
    \centering
    \caption{\label{tab:iqa_cross_domain}IQA cross-domain results: \textit{Ours} vs \textit{Q-align}. Values are SRCC/PLCC, rounded to two decimals. 
    In each cell, the better pair is bolded by the average \((\text{SRCC}+\text{PLCC})/2\).}
    \begin{tabular}{l|c|c|c}
        \toprule
        \backslashbox{Train}{Eval} & KonIQ & SPAQ & KADID \\
        \midrule
        KonIQ (ours)          & \textbf{0.942/0.952} & \textbf{0.899/0.897} & 0.658/0.659 \\
        \hspace{1em}Q-Align \cite{wu2024qalign} & 0.940/0.941 & 0.887/0.886 & \textbf{0.684/0.674} \\
        \midrule
        SPAQ (ours)          & 0.843/0.874 & 0.926/0.930 & 0.677/0.685 \\
        \hspace{1em}Q-Align \cite{wu2024qalign} & \textbf{0.848/0.879} & \textbf{0.930/0.933} & \textbf{0.743/0.740} \\
        \midrule
        KADID (ours)       & 0.639/0.657 & \textbf{0.866/0.865} & \textbf{0.978/0.981} \\
        \hspace{1em}Q-Align \cite{wu2024qalign} & \textbf{0.668/0.665} & 0.860/0.854 & 0.919/0.918 \\
        \bottomrule
    \end{tabular}
\end{table}

\textbf{Comparison with existing methods}. Comparing with established baselines (\cref{tab:ava_perf}, rows 3-8; \cref{tab:iqa_perf_image_only}), image-only RvTC achieves state-of-the-art performance on AVA with SRCC of 0.833 and PLCC of 0.831, surpassing the previous best method One-Align by 1.0 and 1.2 correlation points respectively. 
On IQA datasets, RvTC matches or exceeds existing state-of-the-art across all benchmarks: achieving equivalent performance to Q-Align on \mbox{KonIQ-10k} and SPAQ, while substantially outperforming all methods on KADID-10k. 
Furthermore, in the cross-domain evaluation (\cref{tab:iqa_cross_domain}), RvTC and Q-align achieve broadly similar performance, with each method holding slight advantages in different train–test settings, indicating comparable generalization ability across datasets.

\textbf{Ineffectiveness of current multimodal strategies}. \cref{tab:ava_no_vocab} reveals a key finding that challenges current assumptions about MLLM fine-tuning for regression. Neither human-based vocabulary constraints nor generic task-level prompts improve performance over image-only training. Fine-tuning mPLUG-Owl2 with its original vocabulary classification head and generic prompts yields nearly identical performance to our image-only RvTC model with 5 bins, the number of words used in the syllabus of Q-Align. This confirms that current multimodal approaches fail to leverage cross-modal understanding.

\begin{table}
    \centering
    \caption{\label{tab:ava_no_vocab}
      The importance of task-level prompts and the model's vocabulary. Results on AVA.
      }
    \begin{tabular}{lll}
        \toprule
        Method & SRCC & PLCC \\ 
        \midrule
        Q-Align & 0.822 & 0.817 \\
        \hdashline
        Q-Align (reproduced) & 0.821 & 0.818 \\
        Reversed Syllabus & 0.821 & 0.818 \\
        Alternative Syllabus & 0.821 & 0.819 \\
        Image-Only & 0.823 & 0.820 \\
        \hdashline
        RvTC - Image-Only (5 bins) & 0.823 & 0.818 \\
        RvTC - Image-Only (51 bins) & \textbf{0.833} & \textbf{0.831} \\
        \bottomrule
    \end{tabular}
\end{table}

\textbf{Impact of regression formulation}. 
The progression from vocabulary-constrained classification to unconstrained bin-classification shows clear benefits. 
Moving from \mbox{Q-Align's} 5-token vocabulary approach to RvTC with 5 bins yields similar performance (SRCC 0.823), but increasing to 51 bins provides a substantial boost to 0.833. 
Note that both "Reversed Syllabus" (which inverts Q-Align's original vocabulary order) and "Alternative Syllabus" (using a completely different 5-token vocabulary: "fish", "round", "dash", "car", "end") achieve nearly identical performance, verifying that results are independent of specific vocabulary choices.
This demonstrates that discretization granularity through increased bin count is more effective than attempting to align outputs with human vocabulary patterns, eliminating the need for manual vocabulary crafting while achieving superior performance.

\textbf{Implications for multimodal approaches}. 
These findings establish that current multimodal fine-tuning strategies provide no benefit over carefully designed image-only training. 
\cref{fig:grid_rvtc}'s bin analysis shows that our framework offers a simple yet powerful alternative to complex distributional modeling approaches.
However, this raises the question of whether multimodal capabilities provide additional value, a question we address in \cref{sec:group_level_prompts}.

\subsection{Impact of data-specific prompts}
\label{sec:group_level_prompts}
In this section, we investigate whether semantically meaningful prompts can unlock the cross-modal capabilities of MLLMs. 
We explore this by fine-tuning RvTC on the AVA dataset while incorporating challenge titles as data-specific prompts for each image.

\textbf{Challenge titles as semantic descriptors}. The AVA dataset contains rich semantic information in the form of challenge titles that characterize images with descriptive phrases. Examples include "Rule of Thirds", "School Days Geometry", "Shoes" and "Stationary". These titles are concise yet semantically relevant, providing meaningful context that relates directly to the visual content and aesthetic properties being assessed.

\textbf{Robustness to prompt formulation}. To establish the role of prompt design methodology in our approach, we conducted experiments examining the sensitivity of RvTC to prompt template variations. Our analysis on AVA reveals that adding challenge titles without any additional text produces identical results to elaborate prompt templates that incorporate the challenge title, demonstrating that semantic content rather than template structure drives performance. This finding indicates that RvTC's effectiveness stems from semantic relevance rather than precise prompt engineering, reducing the need for extensive template optimization that typically characterizes other multimodal methods.

\textbf{Performance improvements}. 
\cref{tab:ava_perf} demonstrates that challenge titles substantially improve performance. 
Linear probing (RvTC-LP+) gains 3.3 correlation points (0.709$\rightarrow$0.742 average SRCC and PLCC) without backbone fine-tuning, while full fine-tuning (RvTC+) achieves 0.90 correlation, a 7-point improvement from our already state-of-the-art image-only baseline of 0.83.

\subsection{Ablation studies and analysis}
We conduct ablation studies to understand the mechanisms behind RvTC's performance improvements. 
Through controlled experiments, we demonstrate that incorporating challenge titles yields gains from cross-modal understanding rather than statistical artifacts by decomposing improvements into inter-challenge (statistical bias) and intra-challenge (semantic understanding) components. 
We validate method generalizability across different MLLM architectures and analyze how bin count and data-specific prompts affect performance and training stability.

\textbf{Decomposing performance improvements}.
We distinguish between two types of performance improvements:
\textit{Inter-challenge improvement} refers to gains driven by differences in the statistical properties of challenge-specific data subsets. These improvements are reflected in per-challenge average predicted Mean Opinion Scores (MOS) and could potentially be achieved through statistical pattern recognition without semantic understanding.
\textit{Intra-challenge improvement}, conversely, is measured by correlation metrics of predictions within individual challenges. This type of improvement cannot be directly explained by per-challenge statistical biases and indicates the model's ability to capture meaningful cross-modal relationships between textual and visual features.

\begin{figure}
\begin{center}
\centerline{\includegraphics[width=\columnwidth]{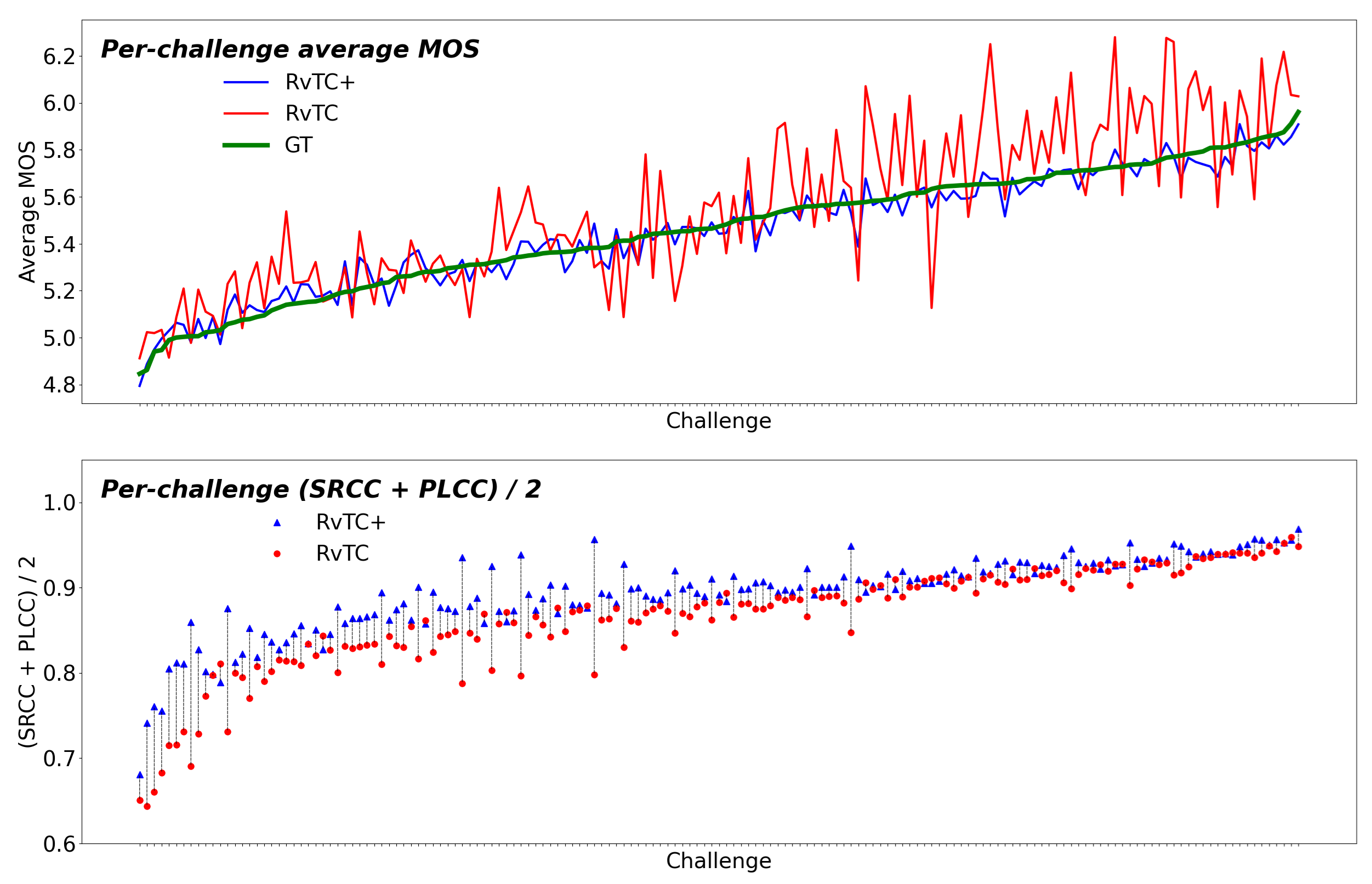}}
\caption{\label{fig:per_challenge30}Performance analysis of RvTC on AVA, comparing image-only RvTC (red) and incorporating image titles RvTC+ (blue) per-challenge average MOS predictions (top figure) and intra-challenge correlation of predictions (bottom figure) implying that the model is leveraging cross-modal features}
\end{center}
\end{figure}

\textbf{Experimental design}.
To isolate these two sources of improvement, we conduct controlled ablation studies comparing image-only RvTC with RvTC+ across different prompt configurations (\cref{tab:ava_ctx_ablation} and \cref{fig:per_challenge30}; see \cref{tab:agi} for comparable analysis on AGIQA-3k).

In \cref{tab:ava_ctx_ablation}, "RvTC - Challenge ID" we replace each challenge title in the textual prompt with a unique challenge identifier (positive integer string). 
This configuration allows the model to leverage inter-challenge statistical bias while eliminating semantic content.

In "RvTC - Shuffled Titles" we randomly reassign challenge titles such that each challenge corresponds to a different title throughout training and evaluation. We ensure no challenge retains its original title. This disrupts semantic coherence while preserving the grouping effect that enables statistical bias exploitation.

In "RvTC+ (with challenge titles)" we provide the full textual challenge title enabling exploitation of both statistical biases and rich semantic cues.

\begin{table}
    \centering
    \caption{\label{tab:ava_ctx_ablation}Ablation study analyzing whether improvements of RvTC on AVA using challenge titles stem from inter-challenge statistical biases by removing the semantic information from the challenge title during fine-tuning}    
    \begin{tabular}{lcll}
        \toprule
        Method & SRCC & PLCC \\ 
        \midrule
        RvTC (image-only) & 0.833 & 0.831 \\
        RvTC - Challenge ID & 0.851 & 0.843 \\
        RvTC - Shuffled Titles & 0.860 & 0.851 \\
        RvTC+ (with challenge titles) & \textbf{0.899} & \textbf{0.901} \\
        \bottomrule
    \end{tabular}
\end{table}

\textbf{Key findings}.
\cref{tab:ava_ctx_ablation} reveals that the overall improvement from incorporating challenge titles cannot be explained merely by inter-challenge statistical biases. The progression from image-only RvTC to full challenge titles demonstrates substantial gains beyond what statistical artifacts can account for.

\cref{fig:per_challenge30} complements these findings. The analysis compares image-only RvTC (red) and RvTC+ (blue) performance within individual challenges containing at least 30 test images. The results demonstrate:
(1) Enhanced per-challenge average MOS predictions (top): Improvements that can be partially attributed to inter-challenge statistical bias exploitation.
(2) Improved intra-challenge correlation of predictions (bottom): Gains that indicate cross-modal feature utilization.

The substantial improvement in intra-challenge correlations provides compelling evidence that the model leverages semantic understanding rather than merely exploiting statistical patterns.

\textbf{Implications}.
These findings establish that incorporating semantically meaningful challenge titles enables RvTC to access and utilize cross-modal understanding capabilities that are not utilized when training with generic prompts or image-only inputs.
The decomposition analysis confirms that a significant portion of the observed performance gains stems from multimodal reasoning rather than statistical bias exploitation, validating the importance of semantic coherence in multimodal regression tasks.

\begin{table}
    \centering
    \caption{\label{tab:rvtc_cross_model}Results across different backbones on the AVA dataset}
    \begin{tabular}{l|cc|cc}
        \toprule
        \multirow{2}{*}{Base Model} & \multicolumn{2}{c|}{RvTC} & \multicolumn{2}{c}{RvTC+} \\
                                   & SRCC & PLCC & SRCC & PLCC \\
        \midrule
        mPLUG-Owl2                & 0.833 & 0.831 & 0.899 & 0.901 \\
        QWEN2-VL-2B              & 0.843 & 0.842 & 0.906 & 0.908 \\
        \bottomrule
    \end{tabular}
\end{table}

\textbf{Backbone generalization}.
To address the generalizability of our RvTC method across MLLMs, we reproduce our experiments on AVA using QWEN2-VL-2B \cite{wang2024qwen2}, a recent vision-language model with different architectural characteristics than mPLUG-Owl2. 
We use identical training procedures and hyperparameters as described in Section 4.1. 
The improvement from RvTC to RvTC+ on both models suggests that our method captures fundamental aspects of vision-language reasoning rather than exploiting model-specific architectural biases. 
This validates that data-specific prompts enable effective cross-modal understanding in MLLMs in fine-tuning for image-based regression tasks.

\begin{figure}[t]
  \centering
  \begin{subfigure}{0.45\textwidth}
    \centering
    \includegraphics[width=\linewidth]{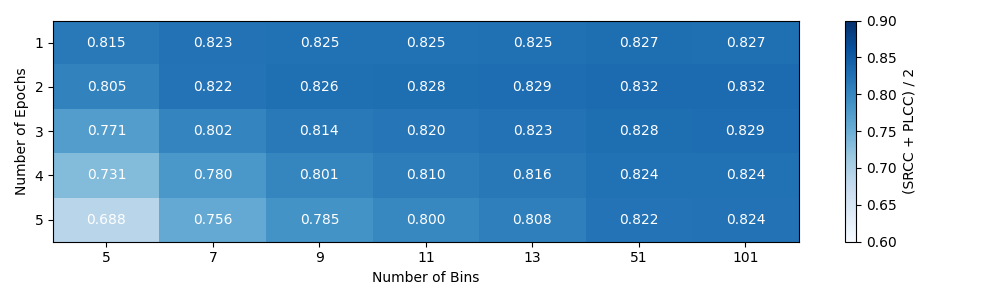}
  \end{subfigure}%
  \hfill
  \begin{subfigure}{0.45\textwidth}
    \centering
    \includegraphics[width=\linewidth]{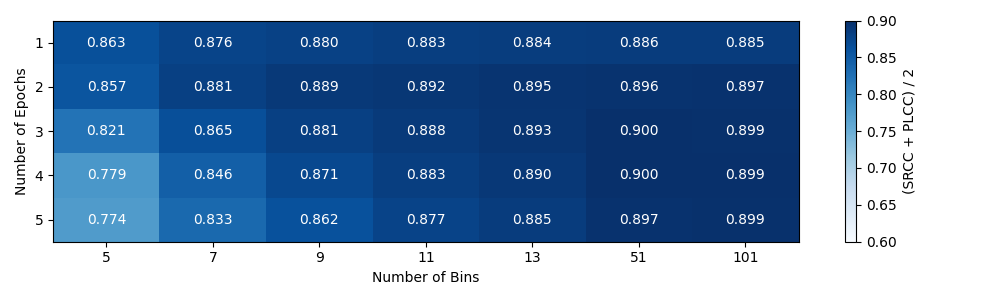}
  \end{subfigure}
  \caption{\label{fig:grid_rvtc}Performance of image-only RvTC (top) and RvTC+ with challenge titles (bottom) on AVA using different number of bins and training lengths}
\end{figure}

\textbf{Bin count analysis}. \cref{fig:grid_rvtc} demonstrates that performance scales monotonically with bin count across different training lengths, with improvement saturating at approximately 51 bins. While increasing bins consistently improves correlation scores, longer training degrades performance in image-only settings due to overconfidence in binning classification.

However, data-specific prompts change this dynamic. \cref{fig:grid_rvtc} (bottom) shows that challenge titles completely eliminate the negative impact of prolonged training when sufficient bins are used, introducing a stabilizing effect that mitigates overfitting scenarios.

\textbf{Training stability advantage}. This stabilization represents a key practical advantage over vocabulary-constrained approaches. 
While previous methods require careful training duration tuning to avoid degradation, RvTC achieves robust stability through increased bin granularity. 
This combination enables longer optimization without extensive hyperparameter tuning, making RvTC more practical for deployment than existing multimodal regression methods.

\subsection{Generalization to AI-generated images}
\label{sec:sample_level_prompts}
To demonstrate the generalizability of our findings beyond aesthetic assessment, we evaluate RvTC on AI-generated image quality assessment using the AGIQA-3k dataset. 
This analysis serves two purposes: (1) validating that data-specific prompts improve performance across different domains, and (2) showing that semantic understanding, rather than statistical bias, drives the observed improvements.

\textbf{Dataset and task formulation}. AGIQA-3k contains image-prompt pairs with two evaluation tasks: \textit{alignment} (how well the generated image matches the prompt instructions) and \textit{perceptual quality} (visual quality of the generated image). 
Each image-prompt pair receives two mean opinion scores (MOS) corresponding to these tasks. 
The semantic content in prompts is directly relevant to the alignment task but less critical for perceptual quality assessment, providing an ideal testbed for examining when and how textual information contributes to regression performance.

\begin{table*}
\centering
\caption{\label{tab:agi}Performance of RvTC on AGIQA-3k's Alignment and Perceptual tasks with different training and evaluation prompts. Bold indicates the best result on each task}
\resizebox{\textwidth}{!}{
\begin{tabular}{l|cc|cc|cc||cc|cc|cc} 
        \toprule
\multirow{3}{*}{\backslashbox{Train}{Eval}} & \multicolumn{6}{c||}{Alignment Task} & \multicolumn{6}{c}{Perceptual Task}  \\ 
\cline{2-13} & \multicolumn{2}{c|}{With prompt} & \multicolumn{2}{c|}{Image-Only} & \multicolumn{2}{c||}{Shuffled Prompt} & \multicolumn{2}{c|}{With prompt} & \multicolumn{2}{c|}{Image-Only} & \multicolumn{2}{c}{Shuffled Prompt}  \\
 & SRCC & PLCC & SRCC & PLCC & SRCC  & PLCC & SRCC & PLCC & SRCC & PLCC & SRCC  & PLCC \\
        \midrule
With Prompt & \textbf{0.810} & \textbf{0.889} & 0.687 & 0.826 & 0.634 & 0.702 & 0.872 & 0.916 & 0.868 & 0.901 & 0.872 & 0.911 \\
Image-Only  & 0.715 & 0.817 & 0.692 & 0.826 & 0.679 & 0.756 & 0.869 & 0.905 & \textbf{0.878} & \textbf{0.918} & 0.865 & 0.884 \\
Shuffled Prompt & 0.672 & 0.828 & 0.678 & 0.827 & 0.676 & 0.828 & 0.872 & 0.914 & 0.862 & 0.902 & 0.872 & 0.915 \\
        \bottomrule
\end{tabular}
}
\end{table*}

\begin{table}
\centering
\caption{\label{tab:prompt_gating}Prompt-Gated Regression on AGIQA-3k. 
Models fine-tuned on a single task (rows 1 and 2), compared to a model fine-tuned on both tasks (row 3).
All models trained with textual prompts.}
\begin{tabular}{l|ll|ll}
        \toprule
\multirow{2}{*}{\backslashbox{Train}{Eval}} & \multicolumn{2}{l|}{Alignment Task} & \multicolumn{2}{l}{Perception Task} \\ 
 & \multicolumn{1}{l}{SRCC} & \multicolumn{1}{l|}{PLCC} & \multicolumn{1}{l}{SRCC} & \multicolumn{1}{l}{PLCC} \\
        \midrule
Alignment & \textbf{0.810} & \textbf{0.889} & 0.709 & 0.793 \\
Perception  & 0.676 & 0.781 & \textbf{0.872} & \textbf{0.916} \\
{Align. + Percep.} & {0.804} & {0.885} & {0.875} & {0.913} \\
        \bottomrule
\end{tabular}
\end{table}

\textbf{Experimental design for bias analysis}. \cref{tab:agi} systematically compares all training and evaluation combinations across both tasks to isolate semantic understanding from statistical artifacts. 
We evaluate three prompt configurations: (1) \textit{original prompts} that maintain semantic coherence, (2) \textit{shuffled prompts} where each image is randomly paired with a different prompt, disrupting semantic alignment while preserving statistical patterns, and (3) \textit{image-only} that removes textual input entirely.

\textbf{Evidence for semantic understanding}. 
For the alignment task, training and evaluation with original prompts achieves optimal performance. Importantly, when a prompt-trained model is evaluated with shuffled prompts, performance degrades substantially below even the image-only baseline, demonstrating that the model has learned semantic associations rather than statistical patterns.


\textbf{Task-specific prompt sensitivity}. In contrast, the perceptual quality task shows minimal sensitivity to prompt variations. Performance remains largely stable across all prompt conditions (SRCC: 0.872$\pm$0.006), indicating that visual quality assessment relies primarily on image features rather than semantic context. This differential sensitivity validates that prompt utility depends on task semantics: prompts enhance performance when semantically relevant but provide little benefit for purely visual assessment tasks.

Note that when fine-tuning with shuffled prompts, both tasks produce an image-only model that largely ignores input prompts. This confirms that the model learns to disregard textual input when it lacks semantic coherence with the visual content.

\textbf{Multi-task learning through prompt-gated regression}. \cref{tab:prompt_gating} demonstrates multi-task learning capabilities through what we term \textit{prompt-gated regression}. 
By adding task identifiers ("Task: image alignment" or "Task: image perceptual quality") to prompts, a single model can simultaneously learn both regression tasks \textit{on the same data}. 
The unified model achieves performance within 0.06 SRCC points of task-specific models, demonstrating that prompt formatting can effectively gate different regression objectives within a single framework.
This extends previous results in \cite{wu2024qalign} where it was shown that a single regression model can be trained on several datasets simultaneously.

\begin{figure}
\vskip 0.2in 
\begin{center}
\centerline{\includegraphics[width=\columnwidth]{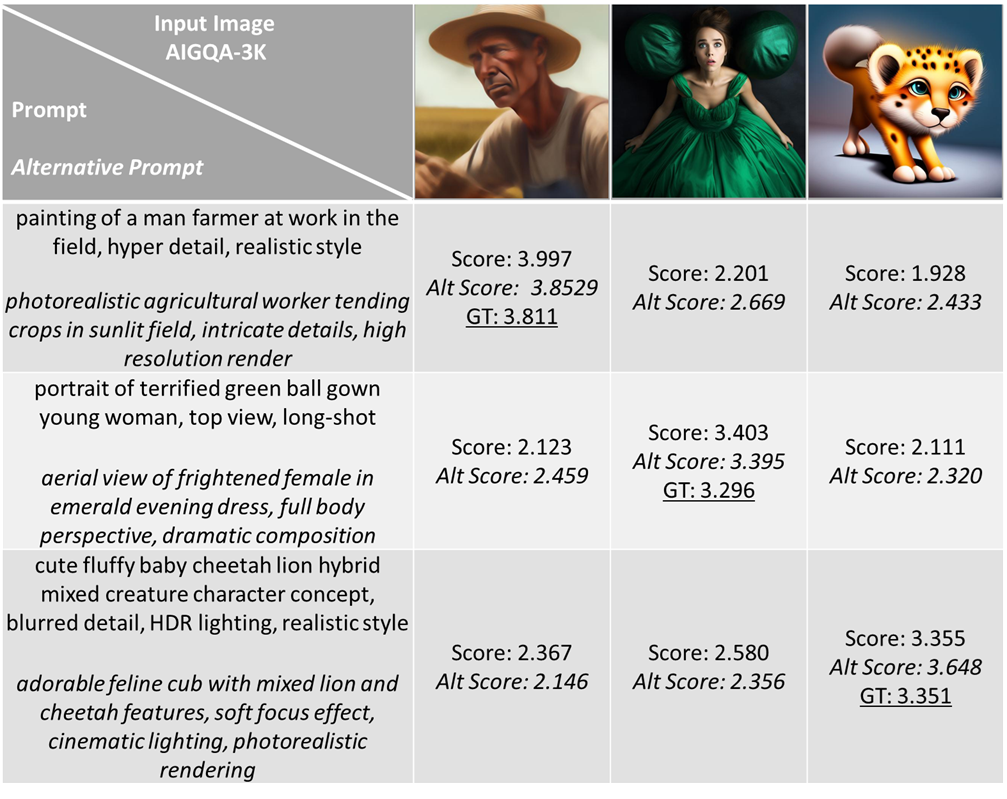}}
\caption{Performance of RvTC fine-tuned on AGIQA-3k when evaluated with original prompt and with \textit{alternative prompt}}
\label{fig:agi_example}
\end{center}
\vskip -0.2in
\end{figure}

\textbf{Robustness to semantic paraphrasing}. To further validate semantic understanding, we test whether the model can generalize to paraphrased prompts that preserve meaning while altering surface form. Using GPT-generated alternative prompts that rephrase original instructions with different structure and vocabulary while maintaining semantic content (\cref{fig:agi_example}), we evaluate whether performance depends on exact phrasing or underlying semantics.

\cref{tab:agi_alt} shows that performance remains stable when using semantically equivalent but syntactically different prompts. For both tasks, correlation scores show minimal variation (alignment: SRCC from 0.810 to 0.809; perceptual: SRCC from 0.872 to 0.874), confirming that the model captures semantic content rather than memorizing specific phrasings. 
This robustness to paraphrasing provides additional evidence that improvements stem from cross-modal understanding.

\textbf{Implications for multimodal regression}. These findings establish three key principles for effective multimodal regression: (1) prompt utility is task-dependent and tied to semantic relevance, (2) statistical bias cannot explain the observed improvements, as evidenced by performance degradation under semantic misalignment, and (3) models can achieve robust semantic understanding that generalizes across different linguistic expressions of the same concepts. Together, these results demonstrate that our approach successfully unlocks cross-modal capabilities in MLLMs for regression tasks when provided with semantically meaningful textual context.

\begin{table}
\centering
\caption{\label{tab:agi_alt}Performance of RvTC fine-tuned on AGIQA-3k with original prompts and evaluated with both original and alternative prompts}
\resizebox{\columnwidth}{!}{
\begin{tabular}{l|ll|ll}
        \toprule
\multirow{2}{*}{\backslashbox{Prompt}{Task}} & \multicolumn{2}{l|}{Alignment Task} & \multicolumn{2}{l}{Perception Task}  \\
 & SRCC & PLCC & SRCC & PLCC \\ 
        \midrule
\begin{tabular}[c]{@{}l@{}}Original Prompts\end{tabular} & 0.810 & 0.889 & 0.872 & 0.916 \\
\begin{tabular}[c]{@{}l@{}}Alternative Prompts\end{tabular} & 0.809 & 0.889 & 0.874 & 0.917 \\
        \bottomrule
\end{tabular}
}
\end{table}

\section{Conclusions}
This work challenges current multimodal large language model approaches to image-based regression tasks. 
Three key insights emerge: First, vocabulary constraints hinder performance; simple bin-based classification outperforms complex vocabulary-dependent methods. 
Second, semantic relevance in prompts is crucial for unlocking cross-modal capabilities, while generic task descriptions provide no benefit over image-only training. 
Third, simple bin-based classification often outperforms complex modeling approaches. Results across different backbones demonstrate that these principles generalize across various MLLM architectures.
This work provides a foundation for more effective multimodal regression systems and underscores the critical role of semantic coherence in cross-modal understanding. 

\textbf{Future work}. Future directions include evaluating RvTC beyond image assessment domains, exploring prompt generation for datasets without textual annotations, investigating whether image-only fine-tuning can preserve multimodal capabilities, and exploring multitask setups combining regression and text generation tasks.

{
    \small
    \bibliographystyle{ieeenat_fullname}
    \bibliography{main}

\begin{thebibliography}{26}
\providecommand{\natexlab}[1]{#1}
\providecommand{\url}[1]{\texttt{#1}}
\expandafter\ifx\csname urlstyle\endcsname\relax
  \providecommand{\doi}[1]{doi: #1}\else
  \providecommand{\doi}{doi: \begingroup \urlstyle{rm}\Url}\fi

\bibitem[Fang et~al.(2020)Fang, Zhu, Zeng, Ma, and Wang]{fang2020spaq}
Yuming Fang, Hanwei Zhu, Yan Zeng, Kede Ma, and Zhou Wang.
\newblock Perceptual quality assessment of smartphone photography.
\newblock In \emph{IEEE Conf. Comput. Vis. Pattern Recog.}, pages 3677--3686, 2020.

\bibitem[Hosu et~al.(2020)Hosu, Lin, Sziranyi, and Saupe]{hosu2020koniq}
Vlad Hosu, Hanhe Lin, Tamas Sziranyi, and Dietmar Saupe.
\newblock Koniq-10k: An ecologically valid database for deep learning of blind image quality assessment.
\newblock \emph{IEEE Trans. Image Process.}, 29:\penalty0 4041--4056, 2020.

\bibitem[Ke et~al.(2021)Ke, Wang, Wang, Milanfar, and Yang]{ke2021musiq}
Junjie Ke, Qifei Wang, Yilin Wang, Peyman Milanfar, and Feng Yang.
\newblock Musiq: Multi-scale image quality transformer.
\newblock In \emph{Int. Conf. Comput. Vis.}, pages 5148--5157, 2021.

\bibitem[Ke et~al.(2023)Ke, Ye, Yu, Wu, Milanfar, and Yang]{ke2023vila}
Junjie Ke, Keren Ye, Jiahui Yu, Yonghui Wu, Peyman Milanfar, and Feng Yang.
\newblock Vila: Learning image aesthetics from user comments with vision-language pretraining.
\newblock In \emph{IEEE Conf. Comput. Vis. Pattern Recog.}, pages 10041--10051, 2023.

\bibitem[Kim et~al.(2021)Kim, Son, and Kim]{kim2021vilt}
Wonjae Kim, Bokyung Son, and Ildoo Kim.
\newblock Vilt: Vision-and-language transformer without convolution or region supervision.
\newblock In \emph{Int. Conf. Mach. Learn.}, pages 5583--5594, 2021.

\bibitem[Kingma and Ba(2014)]{kingma2014adam}
Diederik~P Kingma and Jimmy Ba.
\newblock Adam: A method for stochastic optimization.
\newblock \emph{arXiv preprint arXiv:1412.6980}, 2014.

\bibitem[Lai et~al.(2025{\natexlab{a}})Lai, Chen, Lin, Ye, Liu, Fei, Xing, Wu, Wang, and Zhu]{lai2025snowmaster}
Jianyu Lai, Sixiang Chen, Yunlong Lin, Tian Ye, Yun Liu, Song Fei, Zhaohu Xing, Hongtao Wu, Weiming Wang, and Lei Zhu.
\newblock Snowmaster: Comprehensive real-world image desnowing via mllm with multi-model feedback optimization.
\newblock In \emph{IEEE Conf. Comput. Vis. Pattern Recog.}, pages 4302--4312, 2025{\natexlab{a}}.

\bibitem[Lai et~al.(2025{\natexlab{b}})Lai, Xu, Shi, Yang, Li, Luo, and Li]{lai2025font}
Yingxin Lai, Cuijie Xu, Haitian Shi, Guoqing Yang, Xiaoning Li, Zhiming Luo, and Shaozi Li.
\newblock Font-agent: Enhancing font understanding with large language models.
\newblock In \emph{IEEE Conf. Comput. Vis. Pattern Recog.}, pages 19670--19680, 2025{\natexlab{b}}.

\bibitem[Li et~al.(2023)Li, Zhang, Wu, Sun, Min, Liu, Zhai, and Lin]{li2023agiqa}
Chunyi Li, Zicheng Zhang, Haoning Wu, Wei Sun, Xiongkuo Min, Xiaohong Liu, Guangtao Zhai, and Weisi Lin.
\newblock Agiqa-3k: An open database for ai-generated image quality assessment.
\newblock \emph{IEEE Trans. Circuit Syst. Video Technol.}, 34\penalty0 (8):\penalty0 7465--7481, 2023.

\bibitem[Lin et~al.(2020)Lin, Hosu, and Saupe]{lin2020kadid}
Hanhe Lin, Vlad Hosu, and Dietmar Saupe.
\newblock Deepfl-iqa: Weak supervision for deep iqa feature learning.
\newblock \emph{arXiv preprint arXiv:2001.08113}, 2020.

\bibitem[Liu et~al.(2024)Liu, Li, Li, and Lee]{liu2024llava}
Haotian Liu, Chunyuan Li, Yuheng Li, and Yong~Jae Lee.
\newblock Improved baselines with visual instruction tuning.
\newblock In \emph{IEEE Conf. Comput. Vis. Pattern Recog.}, pages 26296--26306, 2024.

\bibitem[Loshchilov and Hutter(2016)]{loshchilov2016sgdr}
Ilya Loshchilov and Frank Hutter.
\newblock Sgdr: Stochastic gradient descent with warm restarts.
\newblock \emph{arXiv preprint arXiv:1608.03983}, 2016.

\bibitem[Murray et~al.(2012)Murray, Marchesotti, and Perronnin]{murray2012ava}
Naila Murray, Luca Marchesotti, and Florent Perronnin.
\newblock Ava: A large-scale database for aesthetic visual analysis.
\newblock In \emph{IEEE Conf. Comput. Vis. Pattern Recog.}, pages 2408--2415, 2012.

\bibitem[Niu et~al.(2016)Niu, Zhou, Wang, Gao, and Hua]{niu2016ordinal}
Zhenxing Niu, Mo Zhou, Le Wang, Xinbo Gao, and Gang Hua.
\newblock Ordinal regression with multiple output cnn for age estimation.
\newblock In \emph{IEEE Conf. Comput. Vis. Pattern Recog.}, pages 4920--4928, 2016.

\bibitem[Peng et~al.(2024)Peng, Fu, Ming, Wang, Ma, He, Dou, and Chen]{peng2024aigc}
Fei Peng, Huiyuan Fu, Anlong Ming, Chuanming Wang, Huadong Ma, Shuai He, Zifei Dou, and Shu Chen.
\newblock Aigc image quality assessment via image-prompt correspondence.
\newblock In \emph{IEEE Conf. Comput. Vis. Pattern Recog. Worksh.}, pages 6432--6441, 2024.

\bibitem[Radford et~al.(2021)Radford, Kim, Hallacy, Ramesh, Goh, Agarwal, Sastry, Askell, Mishkin, Clark, et~al.]{radford2021clip}
Alec Radford, Jong~Wook Kim, Chris Hallacy, Aditya Ramesh, Gabriel Goh, Sandhini Agarwal, Girish Sastry, Amanda Askell, Pamela Mishkin, Jack Clark, et~al.
\newblock Learning transferable visual models from natural language supervision.
\newblock In \emph{Int. Conf. Mach. Learn.}, pages 8748--8763, 2021.

\bibitem[Talebi and Milanfar(2018)]{talebi2018nima}
Hossein Talebi and Peyman Milanfar.
\newblock Nima: Neural image assessment.
\newblock \emph{IEEE Trans. Image Process.}, 27\penalty0 (8):\penalty0 3998--4011, 2018.

\bibitem[Torgo and Gama(1997)]{torgo1997regression}
Luis Torgo and Joao Gama.
\newblock Regression using classification algorithms.
\newblock \emph{Intelligent Data Analysis}, 1\penalty0 (4):\penalty0 275--292, 1997.

\bibitem[Touvron et~al.(2023)Touvron, Martin, Stone, Albert, Almahairi, Babaei, Bashlykov, Batra, Bhargava, Bhosale, et~al.]{touvron2023llama}
Hugo Touvron, Louis Martin, Kevin Stone, Peter Albert, Amjad Almahairi, Yasmine Babaei, Nikolay Bashlykov, Soumya Batra, Prajjwal Bhargava, Shruti Bhosale, et~al.
\newblock Llama 2: Open foundation and fine-tuned chat models.
\newblock \emph{arXiv preprint arXiv:2307.09288}, 2023.

\bibitem[Wang et~al.(2024)Wang, Bai, Tan, Wang, Fan, Bai, Chen, Liu, Wang, Ge, et~al.]{wang2024qwen2}
Peng Wang, Shuai Bai, Sinan Tan, Shijie Wang, Zhihao Fan, Jinze Bai, Keqin Chen, Xuejing Liu, Jialin Wang, Wenbin Ge, et~al.
\newblock Qwen2-vl: Enhancing vision-language model's perception of the world at any resolution.
\newblock \emph{arXiv preprint arXiv:2409.12191}, 2024.

\bibitem[Wu et~al.(2024{\natexlab{a}})Wu, Zhang, Zhang, Chen, Liao, Wang, Xu, Li, Hou, Zhai, et~al.]{wu2024qinstruct}
Haoning Wu, Zicheng Zhang, Erli Zhang, Chaofeng Chen, Liang Liao, Annan Wang, Kaixin Xu, Chunyi Li, Jingwen Hou, Guangtao Zhai, et~al.
\newblock Q-instruct: Improving low-level visual abilities for multi-modality foundation models.
\newblock In \emph{IEEE Conf. Comput. Vis. Pattern Recog.}, pages 25490--25500, 2024{\natexlab{a}}.

\bibitem[Wu et~al.(2024{\natexlab{b}})Wu, Zhang, Zhang, Chen, Li, Liao, Wang, Zhang, Sun, Yan, Min, Zhai, and Lin]{wu2024qalign}
Haoning Wu, Zicheng Zhang, Weixia Zhang, Chaofeng Chen, Chunyi Li, Liang Liao, Annan Wang, Erli Zhang, Wenxiu Sun, Qiong Yan, Xiongkuo Min, Guangtao Zhai, and Weisi Lin.
\newblock Q-align: Teaching lmms for visual scoring via discrete text-defined levels.
\newblock In \emph{Int. Conf. Mach. Learn.}, pages 54015--54029, 2024{\natexlab{b}}.

\bibitem[Yang et~al.(2024)Yang, Fu, Zhang, Cao, Liu, and Peng]{yang2024moe}
Junfeng Yang, Jing Fu, Wei Zhang, Wenzhi Cao, Limei Liu, and Han Peng.
\newblock Moe-agiqa: Mixture-of-experts boosted visual perception-driven and semantic-aware quality assessment for ai-generated images.
\newblock In \emph{IEEE Conf. Comput. Vis. Pattern Recog. Worksh.}, pages 6395--6404, 2024.

\bibitem[Ye et~al.(2024)Ye, Xu, Ye, Yan, Hu, Liu, Qian, Zhang, and Huang]{ye2024mplug}
Qinghao Ye, Haiyang Xu, Jiabo Ye, Ming Yan, Anwen Hu, Haowei Liu, Qi Qian, Ji Zhang, and Fei Huang.
\newblock mplug-owl2: Revolutionizing multi-modal large language model with modality collaboration.
\newblock In \emph{IEEE Conf. Comput. Vis. Pattern Recog.}, pages 13040--13051, 2024.

\bibitem[You et~al.(2025)You, Cai, Gu, Xue, and Dong]{you2025teaching}
Zhiyuan You, Xin Cai, Jinjin Gu, Tianfan Xue, and Chao Dong.
\newblock Teaching large language models to regress accurate image quality scores using score distribution.
\newblock In \emph{IEEE Conf. Comput. Vis. Pattern Recog.}, pages 14483--14494, 2025.

\bibitem[Zhang et~al.(2023)Zhang, Zhai, Wei, Yang, and Ma]{zhang2023liqe}
Weixia Zhang, Guangtao Zhai, Ying Wei, Xiaokang Yang, and Kede Ma.
\newblock Blind image quality assessment via vision-language correspondence: A multitask learning perspective.
\newblock In \emph{IEEE Conf. Comput. Vis. Pattern Recog.}, pages 14071--14081, 2023.

\end{thebibliography}
}


\end{document}